# Bucket elimination: A unifying framework for probabilistic inference

Rina Dechter
Department of Information and Computer Science
University of California, Irvine
dechter@ics.uci.edu

## Abstract

Probabilistic inference algorithms for finding the most probable explanation, the maximum aposteriori hypothesis, and the maximum expected utility and for updating belief are reformulated as an elimination–type algorithm called *bucket elimination*. This emphasizes the principle common to many of the algorithms appearing in that literature and clarifies their relationship to nonserial dynamic programming algorithms. We also present a general way of combining conditioning and elimination within this framework. Bounds on complexity are given for all the algorithms as a function of the problem's structure.

## 1 INTRODUCTION

An external observer attempting to sort out the core ideas behind current algorithms for processing influence diagrams or Bayesian networks normally find the topic confusing; the variety of paradigm nomenclatures and implementation considerations in the literature is enormous. Some of the ideas are translations of each other, others involve combinations of existing ideas, others are extensions. Yet, the relationships among the various approaches are not always explicitly stated.

Here, I wish to present a purely algorithmic view of the core idea behind the main approach to probabilistic reasoning, in the hope that this view will make the current literature more accessible to newcomers. This view, called *bucket elimination*, is a generalization of nonserial dynamic programming à la Bertele and Briochi [BeBr 72]. It allows a uniform way of combining elimination with conditioning, and provides insight into the relationship between clustering and elimination.

To emphasize the generality of bucket elimination we start with a similar algorithm in the area of deterministic reasoning. Consider the following algorithm for deciding the satisfiability of a propositional theory in *Conjunctive Normal Form (CNF)*. Given a set of clauses and given an ordering of the propositional variables, assign to each clause the index of the highest ordered literal in that clause. Then resolve only clauses having the same index, and only on their highest literal. The result of this restriction is a systematic elimination of literals from the set of clauses that are candidates for future resolution. This algorithm, which we call *directional resolution* (DR), is the core of the well-known Davis-Putnam algorithm for satisfiability [DaPu 60; DeRi 94].

Algorithm DR (see Figure 1) is described using *buckets* partitioning the set of clauses in the theory $\varphi$. We call its output theory, $E_d(\varphi)$, the *directional extension* of $\varphi$. Given an ordering $d = Q_1, ..., Q_n$, all the clauses containing $Q_i$ that do not contain any symbol higher in the ordering are placed in the bucket of $Q_i$, denoted $bucket_i$. The algorithm processes the buckets in a reverse order of $d$. When processing $bucket_i$, it resolves over $Q_i$ all possible pairs of clauses in the bucket and inserts the resolvents into the appropriate lower buckets. It was shown [DeRi 94] that:

**Theorem 1.1 (model generation)**
*Let $\varphi$ be a cnf formula, $d = Q_1, ..., Q_n$ an ordering, and $E_d(\varphi)$ its directional extension. Then, if the extension is not empty, any model of $\varphi$ can be generated in a backtrack-free manner, consulting $E_d(\varphi)$ in the order d as follows: assign to $Q_1$ a truth value that is consistent with clauses in $bucket_1$ (if the bucket is empty, assign $Q_1$ an arbitrary value); after assigning values to $Q_1, ..., Q_{i-1}$, assign a value to $Q_i$ so that together with the previous assignments it will satisfy all clauses in $bucket_i$.*

It was also shown [DeRi 94], that the complexity of DR is exponentially bounded (time and space) in the induced width (also called tree-width) of the *interaction graph* of the theory, where a node is associated with a proposition and an arc connects any two nodes appearing in the same clause.

The collection of belief network algorithms we present next have a lot in common with the resolution procedure above. They all possess similar properties of

---

**directional resolution**
**Input:** A cnf theory $\varphi$, an ordering $d = Q_1, ..., Q_n$,
**Output:** A decision of whether $\varphi$ is satisfiable. If it is, a theory $E_d(\varphi)$, equivalent to $\varphi$; else, an empty directional extension.
1. **Initialize:** Generate an ordered partition of the clauses, $bucket_1, ..., bucket_n$, where $bucket_i$ contains all the clauses whose highest literal is $Q_i$.
2. For $p = n$ to 1 do:
• If $bucket_p$ contains a unit clause, perform only unit resolution. Put each resolvent in the appropriate bucket.
• **else,** resolve each pair $\{(\alpha \vee Q_p), (\beta \vee \neg Q_p)\} \subseteq bucket_p$. If $\gamma = \alpha \vee \beta$ is empty, return $E_d(\varphi) = \emptyset$, the theory is not satisfiable; else, determine the index of $\gamma$ and add it to the appropriate bucket.
3. Return $E_d(\varphi) \Longleftarrow \bigcup_i bucket_i$.

---

Figure 1: Algorithm directional resolution

compiling a theory into a backtrack-free (e.g., greedy) theory and their complexity is dependent on the same graph parameters. The algorithms are, for the most part, not new in the sense that the basic ideas have existed for some time [Pear 88; Spie 86; TaSh 90; Jens 94; Shac 90; Bacc 95; Shac 86; Shac 88; ShCh 91; Shen 92]. What we advocate is a syntactic and uniform exposition emphasizing the algorithm's form as a straightforward elimination algorithm. The main virtue of this presentation, beyond uniformity, is that it facilitates transfer of ideas and techniques across areas of research. In particular, having noted that elimination algorithms and clustering algorithms are very similar [DePe 89], we propose a uniform way for improving such algorithms based on conditioning. We show that the idea of conditioning, which is as universal as that of elimination, can be incorporated and exploited naturally within the elimination framework. This leads to a hybrid algorithm, one trading off time for space [Dech 96].

The work we present here also fits into the framework of [Arnb 85; ArPr 89]. Arnborg presents table-based reductions for various NP-hard graph problems such as the independent set problem, network reliability, vertex cover, graph $k$-colorability, and Hamilton circuits. Our paper as well as [DeBe 95] extends this approach to a different set of problems.

## 2 PRELIMINARIES

A *belief network* (BN) is a concise description of a complete probability distribution. It is defined by a directed acyclic graph over nodes representing random variables, where each variable is annotated with the conditional probability matrices specifying its probability given each value combination of its parent variables. A BN uses the concept of a directed graph.

**Definition 2.1 (graph concepts)** *A directed graph is a pair, $G = \{V, E\}$, where $V = \{X_1, ..., X_n\}$ is a set of elements and $E = \{(X_i, X_j)|X_i, X_j \in V\}$ is the set of edges. If $(X_i, X_j) \in E$, we say that $X_i$ points to $X_j$. For each variable $X_i$, $pa(X_i)$ is the set of variables pointing to $X_i$ in $G$, while the set of child nodes of $X_i$, denoted $ch(X_i)$, comprises the variables that $X_i$ points to. The family of $X_i$, $F_i$, includes $X_i$ and its child variables. A directed graph is acyclic if it has no directed cycles. In an undirected graph, the directions of the arcs are ignored: $(X_i, X_j)$ and $(X_j, X_i)$ are identical. An ordered graph is a pair $(G, d)$ where $G$ is an undirected graph and $d = X_1, ..., X_n$ is an ordering of the nodes. The width of a node in an ordered graph is the number of its neighbors that precede it in the ordering. The width of an ordering $d$, denoted $w(d)$, is the maximum width over all nodes. The induced width of an ordered graph, $w*(d)$, is the width of the induced ordered graph obtained as follows: nodes are processed from last to first; when node $X$ is processed, all its preceding neighbors are connected. The induced width of a graph, $w*$, is the minimal induced width over all its orderings; it is also known as the tree-width. A graph has an induced width $k$ iff it can be embedded into a $k$-tree, in which case it is called a partial $k$-tree [Arnb 85]. A cycle-cutset is a subset of nodes in the graph that, when removed, results in a graph without cycles.*

**Definition 2.2 (belief networks)**
*Let $X = \{X_1, ..., X_n\}$ be a set of random variables over multivalued domains, $D_1, ..., D_n$. A BN is a pair $(G, P)$ where $G$ is a directed acyclic graph and $P = \{P_i\}$. $P_i$ is the conditional probability matrices associated with $X_i$, $P_i = \{P(X_i|pa(X_i))\}$. An assignment $(X_1 = x_1, ..., X_n = x_n)$ can be abbreviated to $x = (x_1, ..., x_n)$. The BN represents a probability distribution over $X$ having the product form*

$$P(x_1, ...., x_n) = \Pi_{i=1}^n P(x_i|x_{pa(X_i)})$$

*$x_S$ denotes the projection of a tuple $x$ over a subset of variables $S$. An evidence set $e$ is an instantiated subset of variables. $A = a$ denotes a partial assignment to a subset of variables in $A$. Whenever no confusion can arise, we abbreviate $pa(X_i)$ by $pa_i$ and $ch(X_i)$ by $ch_i$. If $u$ is a tuple over a subset $X$, then $u_S$ denotes that assignment, restricted to the variables in $S \cap X$. Let $u$ be a tuple over a subset of variables, $S$ denote a subset of variables, and $X_p$ be a variable not in $S$. In the following, we frequently use $(u_S, x_p)$ to denote the tuple $u_S$ appended by a value $x_p$ of $X_p$. We abbreviate $\bar{x}_i = (x_1, ..., x_i)$ and $\bar{x}_i^j = (x_i, x_{i+1}, ..., x_j)$.*

Next, we focus on several fundamental queries, all defined over a belief network $BN$ and given some evidence $e$:

**Definition 2.3 (queries)**

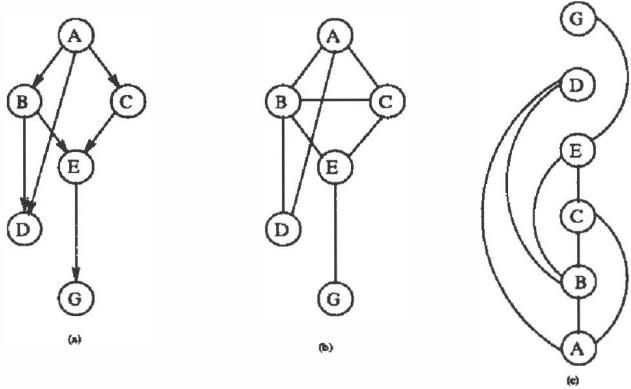

Figure 2: A belief network representing
$P(g, e, d, c, b, a)=$
$P(g|e)P(e|c, b)P(d|b, a)P(b|a)P(c|a)$.

1. **Belief assessment:** *The belief assessment task of $X_i = x_i$ is to find $bel(x_i) = P(X_i = x_i|e)$.*

2. **Most probable explanation (MPE):** *The MPE task is to find an assignment $x^o = (x^o{}_1, ...x^o{}_n)$ such that $p(x^o) = \max_{\bar{x}_n} \Pi_{i=1}^n P(x_i|x_{pa_i}, e)$.*

3. **Maximum aposteriori hypothesis (MAP):** *Given a set of hypothesized variables $A = \{A_1, ...A_k\}$, $A \subseteq X$, the MAP task is to find an assignment $a^o = (a^o{}_1, ...a^o{}_k)$ such that $p(a^o) = \max_{\bar{a}_k} \sum_{x_{X-A}} \Pi_{i=1}^n P(x_i|x_{pa_i}, e)$.*

4. **Maximum expected utility (MEU):** *Given a real-valued utility function $u(x)$, $u(x) \to R$, which is additively decomposable relative to $Q_1, ..., Q_j$, $Q_i \subseteq X$, as follows $u(x) = \sum_{Q_j \in Q} f_j(x_{Q_j})$, and given a subset of decision variables $D = \{D_1, ..., D_k\}$ that are root variables in BN, $D \subseteq X$, the MEU task is to find an assignment $d^o = (d^o{}_1, ..., d^o{}_k)$ such that $(d^o) = argmax_d \sum_{x_{k+1},...,x_n} \Pi_{i=1}^n P(x_i|x_{pa_i,d})u(x)$.*

It is known that these tasks are NP-hard. Nevertheless, a polynomial propagation algorithm for singly-connected networks [Pear 88] exists. The two main approaches to extending this propagation algorithm to multiply-connected networks are the *cycle-cutset* approach, also called *conditioning*, and *tree-clustering* [Pear 88; Spie 86; Shac 86]. These methods work well only for sparse networks with small cycle-cutsets or small clusters. Complexity is time exponential in the cutset size for the former, time and space exponential in the cluster sizes, bounded by the induced-width, for the latter.

## 3 DYNAMIC PROGRAMMING

We now present elimination algorithms for the various tasks. The algorithms generalize the family of nonserial dynamic programming [BeBr 72]. Because dynamic programming algorithms work by eliminating variables one by one while computing the effect of each eliminated variable on the remainder of the problem, they can be viewed as *elimination algorithms*. It is known that most such algorithms have worst-case complexity bounded exponentially by the induced width [Dech 90; Arnb 85] of their *underlying graph*. In belief networks the graph, often called the *moral graph*, is obtained by connecting all the parents of each node in the acyclic graph and ignoring directionality. When the graph is a tree, the elimination algorithms largely coincide with the linear propagation algorithms for trees. Various elimination-type algorithms for processing influence diagrams and BN have been studied [Shac 86; Shac 88; TaSh 90; Shac 90; Jens 94; ShPe 92; Shen 92]. Lack of space prevents us from showing explicitly how they map into the elimination framework.

### 3.1 AN ELIMINATION ALGORITHM FOR MPE

Following Pearl's propagation algorithm for singly-connected networks [Pear 88], researchers have investigated various approaches to finding the MPE in BN. Early attempts are given in [Coop 84; PeRe 86; PeRe 89]. Recent proposals include best first-search algorithms [ShCh 91] and algorithms based on linear programming [Sant 91].

The problem is to maximize the function $\max_x P(x) = \max_x \Pi_i P(x_i|x_{pa_i})$ when $x = (x_1, ..., x_n)$. Consider an arbitrary ordering of the variables $(X_1, ..., X_n)$. Partition the conditional probability matrices $\{P_i\}$ into buckets. In the bucket of $X_i$ put all the matrices mentioning $X_i$ that do not mention any variable higher in the ordering. The procedure has backward and forward parts and is justified by the following symbolic manipulation (see also [Shac 90]).

(1) *Backward part*. Consider variable $X_n$ first (remember $\bar{x}_i = (x_1, ..., x_i)$),

$$M = \max_{\bar{x}_n} P(x) = \max_{\bar{x}_{n-1}} \max_{x_n} \Pi_{i=1}^n P(x_i|x_{pa_i})$$

All the expressions that do not mention $X_n$ can be migrated to the left of the maximization on $X_n$ since, relative to $X_n$, they are constants. The only matrices mentioning $X_n$ are those relating to its Markov neighborhood: its parents, children, and children's parents. Let $U_n$ be the set of all the variables mentioned in the bucket of $X_n$, excluding $X_n$. Initially, prior to processing, this set coincides with $U_n = pa_n \cup ch_n \cup_j F_{nj} - X_n$, where $F_{nj}$ are the parents of $X_n$'s $j^{th}$ child node. We get (Remember that $F_i$ includes $X_i$ and $ch(X_i)$).

$$M = \max_{\bar{x}_{n-1}} \Pi_{\{X_i \in X - F_n\}} P(x_i|x_{pa_i}) \cdot$$

$$\max_{x_n} \Pi_{X_i \in F_n} P(x_i|x_{pa_i}) =$$

$$\max_{\bar{x}_{n-1}} \Pi_{\{X_i \in X - F_n\}} P(x_i|x_{pa_i}) \cdot h_n(x_{U_n})$$

Hence, the first step consists of processing the conditional probability matrix in the bucket of $X_n$ and computing the function $h_n : U_n \rightarrow R$, $h_n(x) = \max_{x_n} \Pi_{X_i \in F_n} P(x_i|x_{pa_i})$. The new function, $h_n$, is placed in the bucket of the largest-index variable amongst $U_n$. The optimizing value of $X_n$ for each tuple $x$ is defined by $x_n^{opt}(x) = argmax_{X_n} h_n(x)$. The procedure continues recursively with the next variable. During processing, the functional components in each bucket are either the original conditional probabilities or functions computed when processing earlier buckets. We will denote all such functions (also called matrices) in each bucket uniformly as $h_1, ..., h_k$ and the variable subsets on which they are defined as $S_1, ..., S_k$.

(2) *Forward part* (processing variable $X_i$ after selecting the partial assignment $x = (x_1, ..., x_{i-1})$). Choose value $x_i^{opt}(x_{U_i})$ recorded in the backward phase.

The algorithm is described in Figure 3. We will demonstrate the backward elimination part of the algorithm using the example in Figure 2. We will assume no evidence for now.

**Example 3.1** *Consider the variables in the order $A, C, B, E, D, G$. Process the variables from last to first and partition the conditional probability matrices into buckets, yielding $bucket_G = \{P(G|E)\}$, $bucket_E = \{P(E|B,C)\}$, $bucket_D = \{P(D|B,A)\}$, and $bucket_B = \{P(B|A)\}$. First, eliminate variable $G$, by computing the maximum probability extension to $G$ of each value of $E$, namely, $h_G(e) = \max_{g \in G} P(e|g)$, and place $h_G(e)$ in $bucket_E$. Then, record the maximizing values $G^{opt}(e) = argmax h_G(e)$ and place the result in $bucket_G$. Subsequently, process $bucket_D$. To eliminate $D$, compute $h_D(b,a) = \max_{d \in D} P(b,a|d)$, place the result in $bucket_B$, and record the values $e^{opt}(b,a)$. Next, process variable $E$. Its bucket now contains two matrices: $P(E|B,C)$ and $h_G(E)$. To eliminate $E$, compute $h_E(b,c) = \max_{e \in E} p(e|b,c) \cdot h_G(e)$ and place the resulting function in $bucket_B$. To eliminate $B$, compute and record the function $h_B(a,c) = \max_{b \in B} P(b|a) \cdot h_D(b,a) \cdot h_E(b,c)$, placing it in $bucket_C$. To eliminate $C$, compute $h_C(a) = \max_{c \in C} P(c|a) \cdot h_B(a,c)$. Finally, compute the maximum value associated with $A$ by computing $h_{max} = \max_{a \in A} h_B(a) \cdot h_C(a)$.*

This backward process can be viewed as a compilation (or learning) phase, in which we compile information that allows the most probable tuple to be generated later without searching or backtracking. We generate the most probable tuple by following the pointers in the recorded tables. In Example 3.1, we recorded two-dimensional functions at the most, and therefore the complexity is at most time and space cubic in the domain sizes.

### 3.1.1 Handling Observations

Given evidence $e$, we will compute the most likely tuple that maximizes the joint probability when the observed variables are assigned their values in $e$. Namely, we compute $\max_x P(x \wedge e)$. The same tuple will maximize also the probability function conditioned on $e$, since those two functions are related by the normalization constant $P(e)$. To accomplish that within the elimination scheme, observed variables are handled by putting each observation in its corresponding bucket. Continuing with our example, suppose we wish to compute the MPE having observed $B = 1$. This observation will have an effect only when processing $bucket_B$. When the algorithm arrives at that bucket, it contains the three matrices $P(b|a)$, $h_D(b,a)$, and $h_E(b,c)$, as well as the observation $B = 1$. According to the processing rule, we will compute, had we not had special case-handling for observations, $h_B(a,c) = P(b = 1|a)h_D(b = 1,a)h_E(b = 1,c)$. Namely, we will generate a two-dimensional function. This is unnecessary, however. It would be more effective to apply the assignment $B = 1$ to each matrix separately and put the resulting functions into buckets separately. In this case we generate $P(b = 1|a)$ and $h_D(b = 1, a)$ which will be placed in the bucket of $A$, and $h_E(b = 1, c)$ that will be placed in the bucket of $C$. We thus avoid increasing the dimensionality of recorded functions. Processing buckets containing observations in this manner exploits the cutset effect of conditioning automatically [Pear 88].

Another important point is that, had the bucket of $B$ been at the top of our ordering, the advantage of this observation could have been exploited earlier in the computation. For example, if we use the ordering $A, C, E, G, D, B$, then we start by processing $bucket_B$ containing $P(b|a), P(d|b,a), P(e|c,b), B = 1$. The special rule for processing buckets holding observations will place $P(b = 1|a)$ in $bucket_A$, $P(d|b = 1, a)$ in $bucket_D$, and $P(e|c,b = 1)$ in $bucket_E$. In subsequent processing, only one-dimensional functions will be recorded, as if the underlying graph is a tree. Consequently, to have the full computational benefit of observations, we may assume that observed variables are placed last in the ordering and therefore, processed first.

### 3.1.2 Complexity

The complexity of algorithm elim-max is bounded by the time and space needed to process a bucket, which is bounded exponentially by the number of variables mentioned in a bucket. It is possible to show, by graph manipulation only, that the maximum number of variables in the bucket of $X_i$ along ordering $d$ is bounded by $w_d^*(X_i)$, the induced width of $X_i$. For instance, the moral graph of the $DAG$ in Figure 2a, is depicted in Figure 2b, the induced graph relative to $d = A, B, C, E, D, G$ is depicted in 2c. The induced width of that ordering (which equals the width in this case) is 2, and, indeed, the maximum arity of functions recorded by elim-max is also 2. The induced width of the reversed ordering is 3, and so is the recorded function's dimensionality. We conclude:

---

**Algorithm elim-max**
**Input:** A belief network $BN = \{P_1, ..., P_n\}$; an ordering of the variables, $o$; observations $e$.
**Output:** The most probable assignment.
1. **Initialize:** Generate an ordered partition of the conditional probability matrices, $bucket_1, ..., bucket_n$, where $bucket_i$ contains all matrices whose highest variable is $X_i$. Put each observed variable in its appropriate bucket. Let $S_1, ..., S_j$ be the subset of variables in the processed bucket, on which matrices (new or old) are defined.
2. **Backward:** for $p \leftarrow n$ downto 1 do
for all the matrices $h_1, h_2, ..., h_j$ in $bucket_p$ do
• If (bucket with observed variable) $bucket_p$ contains $X_p = x_p$, assign $X_p = x_p$ to each $h_i$ and put each in appropriate bucket.
• else, $U_p \leftarrow \bigcup_{i=1}^j S_i - \{X_p\}$. For all $U_p = u$, $h_p(u) = \max_{x_p} \Pi_{i=1}^j h_i(x_p, u_{S_i})$. $x_p^{opt}(u) = argmax_{X_p} h_p(u)$.
Add $h_p$ to bucket of largest-index variable in $U_p$.
3. **Forward:** Assign values in the ordering $o$ using the recorded functions $x^{opt}$ in each bucket.

Figure 3: Algorithm elim-max

**Theorem 3.2** *Given a belief network having n variables, algorithm elim-max is guaranteed to solve the MPE task. The complexity of the algorithm is time and space exponentially bounded in the induced width of the network's ordered moral graph, $O(n \cdot exp(w*(d)))$.* □

### 3.2 AN ELIMINATION ALGORITHM FOR BELIEF ASSESSMENT

The algorithm for belief assessment is identical to elim-max with one change: maximization is replaced by summation. Let $X_1 = x_1$ be an atomic proposition. The problem is to assess and later update the belief in $x_1$ given evidence $e$. Namely, to compute

$$P(X_1 = x_1|e) = \sum_{x=\bar{x}_2^n} \Pi_{i=1}^n P(x_i|x_{pa_i}, e)$$

Consider an ordering of the variables $(X_1, ..., X_n)$. Partition the conditional probability matrices as before. The procedure has only a backward phase. Consider variable $X_n$ first.

$$P(x_1|e) = \sum_{x=\bar{x}_2^n} P(\bar{x}_n|e) = \sum_{\bar{x}_2^{(n-1)}} \sum_{x_n} \Pi_i P(x_i|x_{pa_i}, e) =$$

$$\sum_{x=\bar{x}_2^{(n-1)}} \Pi_{X_i \in X - F_n} P(x_i|x_{pa_i}, e) \cdot$$

$$\sum_{x_n} P(x_n|x_{pa_n}, e)) \Pi_{X_i \in ch_n} P(x_i|x_{pa_i}, e) =$$

$$\sum_{x=\bar{x}_2^{(n-1)}} \Pi_{X_i \in X - F_n} P(x_i|x_{pa_i}, e) \cdot \lambda_n(x_{U_n})$$

**Algorithm elim-bel**
**Input:** A belief network $BN = \{P_1, ..., P_n\}$, and an ordering of the variables, $o = X_1, ..., X_n$.
**Output:** the belief of $X_1$ given evidence $e$.
1. **Initialize:** generate an ordered partition of the conditional probability matrices into buckets. $bucket_i$ contains all matrices whose highest variable is $X_i$. Put each observation in its bucket. Let $S_1, ..., S_j$ be the subset of variables in the processed bucket on which matrices (new or old) are defined.
2. **Backwards:** for $p \leftarrow n$ downto 1 do
for all the matrices $\lambda_1, \lambda_2, ..., \lambda_j$ in $bucket_p$ do
• If ( bucket with observed variable) $X_p = x_p$ appear in bucket, then substitute $X_p = x_p$ in each matrix $\lambda_i$ and put each in appropriate bucket.
• else, $U_p \leftarrow \bigcup_{i=1}^j S_i - \{X_p\}$ For all $U_p = u$, $\lambda_p(u) = \sum_{x_p} \Pi_{i=1}^j \lambda_i(x_p, u_{S_i})$.
Add $\lambda_p$ to the largest index variable in $U_p$.
3. **Return** $Bel(x_1) = \alpha P(x_1) \cdot \Pi_i \lambda_i(x_1)$
(where the $\lambda_i$ are in $bucket_1$, $\alpha$ is a normalizing constant.)

Figure 4: Algorithm elim-bel

Processing $bucket_n$ amounts to computing the function $\lambda_n$. Therefore, when processing each bucket we multiply all the bucket's matrices, $\lambda_1, ..., \lambda_j$, defined over subsets $S_1, ..., S_j$, and then eliminate the bucket's variable by summation. The computed function is $\lambda_n : U_n \rightarrow R$, $\lambda_n(u) = \sum_{x_n} \Pi_{i=1}^j \lambda_i(x_n, u_{S_i})$, where $U_n = \cup_i S_i - X_n$. As before, the computed function is placed in the bucket of its largest-index variable in $U_n$. The procedure continues recursively, processing the bucket of the next variable. After all the buckets are processed, the answer is available in the first bucket. Algorithm elim-bel is described in Figure 4. Observed variables are handled as before.

**Example 3.3** *Consider again the variables in the order $A, C, B, E, D, G$, and assume evidence that $G = 1$. Process variables from last to the first and partition the conditional probability matrices into buckets, getting $bucket_G = \{P(G|E), G = 1\}$, $bucket_E = \{P(E|B,C)\}$, $bucket_D = \{P(D|B,A)\}$, $bucket_B = \{P(B|A)\}$, $bucket_C = \{P(C|A)\}$, and $bucket_A = \{P(A)\}$. To process $G$, assign $G = 1$, get $\lambda_G(e) = P(g = 1|e)$, and place the result in $bucket_E$. Subsequently, process $bucket_D$ by computing $\lambda_D(b,a) = \sum_{d \in D} P(d|b,a)$ and putting the result in $bucket_B$. The bucket of $E$, to be processed next, now contains two matrices: $P(E|B,C)$ and $\lambda_G(E)$. Compute $\lambda_E(b,c) = \sum_{e \in E} p(e|b,c) \cdot \lambda_G(e)$, and place the resulting function in $bucket_B$. To eliminate $B$ we record the function $\lambda_B(a,c) = \sum_{b \in B} P(b|a) \cdot \lambda_D(b,a) \cdot \lambda_E(b,c)$, placing it in $bucket_C$, and to eliminate $C$ we compute $\lambda_C(a) = \sum_{c \in C} P(c|a) \cdot \lambda_B(a,c)$. Finally, in $bucket_A$, we compute the belief in $A = a$, to be $\alpha \cdot P(a) \cdot \lambda_B(a) \cdot \lambda_C(a)$, when $\alpha$ is a normalization constant.*

As before, the complexity of *elim − bel* is bounded exponentially by the dimension of the recorded matrices, which in turns, can be bounded by the induced width of the moral graph relative to the elimination ordering. In summary,

**Theorem 3.4** *Algorithm elim-bel computes the belief of $X_1$. Its complexity is $O(n \cdot exp(w*(d)))$, when $w*(d)$ is the induced width along d of its moral graph, where n is the number of variables.* □

## 3.3   AN ELIMINATION ALGORITHM FOR MAP

We next present an elimination algorithm for the *MAP* task. To simplify exposition we assume that everything is conditioned on subset of observations without explicitly mentioning it. The algorithm is a combination of the prior two; some of the variables are eliminated by summation, others by maximization.

Given a belief network, a subset of hypothesis variables $A = \{A_1, ..., A_k\}$ and some evidence, the problem is to find an assignment to the hypothesized variable that maximizes their probability. Formally we wish to compute $\max_{\bar{a}_k} P(\bar{a}_k) = \max_{\bar{a}_k} \sum_{\bar{x}_{k+1}^n} \Pi_{i=1}^n P(x_i|x_{pa_i})$ when $x = (a_1, ..., a_k, x_{k+1}, ..., x_n)$. When manipulating this expression we can push the maximization to the left of the summation. This means that in the elimination algorithm the maximized variables should initiate the ordering (they would be eliminated last). Therefore, orderings that optimize elimination over $X - A$ should be considered independently of orderings of the summation variables. In algorithm *elim − map* in Figure 5, we will consider only orderings in which the hypothesized variables appear before the rest. The algorithm has a backward phase and its forward phase is only relative to the hypothesis variables. Maximization and summation can be somewhat interleaved allowing more effective orderings. We do not incorportae this option here.

**Theorem 3.5** *Algorithm elim-map computes the MAP task. Its complexity is $O(n \cdot exp(w*(d)))$, when $w*(d)$ is the induced width along d of its moral graph where n is the number of variables.* □

## 3.4   AN ELIMINATION ALGORITHM FOR MEU

The last and most complicated task is to determine a set of decisions that maximize the expected utility, defined on the network. Given a Belief network $BN$, evidence $e$, and a real-valued utility function $u(x)$, $u(x) \to R$, additively decomposable relative to $Q = \{Q_1, ..., Q_j\}$, $Q_i \subseteq X$, and defined by $u(x) = \sum_{Q_j \in Q} f_j(x_{Q_j})$, and given a subset of decision variables $D = \{D_1, ...D_k\}$ which are root nodes, the MEU task is to find a set of decisions $d^o = (d^o{}_1, ..., d^o{}_k)$ that maximizes the expected utility. We assume that the variables not in $D$ are indexed $X_{k+1}, ..., X_n$.

---

**Algorithm elim-map**
**Input:** A belief network $BN = \{P_1, ..., P_n\}$, a subset of variables $A = \{A_1, ..., A_k\}$ and an ordering of the variables, $o$ in which the $A$'s are first in the ordering.
**Output:** A most probable assignment $A = a$.
1. **Initialize:** generate an ordered partition of the conditional probability matrices into $bucket_1$, ..., $bucket_n$, where $bucket_i$ contains all matrices whose highest variable is $X_i$.
2. **Backwards:** for $p \leftarrow n$ downto 1 do
for all the matrices $\beta_1, \beta_2, ..., \beta_j$ in $bucket_p$ do
• **If** (bucket with observed variable) $bucket_p$ contains the observation $X_p = x_p$, then assign $X_p = x_p$ to each $\beta_i$ and put in appropriate bucket.
• else, $U_p \leftarrow \bigcup_{i=1}^{j} S_i - \{X_p\}$. If $X_p$ is not a member of $A$ then, For all $U_p = u$, $\beta_p(u) = \sum_{x_p} \Pi_{i=1}^{j} \beta_i(x_p, u_{S_i})$,
else, $(X_p \in A)$ $\beta_p(u) = \max_{x_p} \Pi_{i=1}^{j} \beta_i(x_p, u_{S_i})$ and $a^0(u) = argmax_{x_p} \beta_p(u)$. Add $\beta_p$ to the bucket of the largest-index variable in $U_p$.
3. **Forward:** Assign values, in the ordering $o = A_1, ..., A_k$ using the information recorded in each bucket.

Figure 5: Algorithm elim-map

Formally, we want to maximize the function (while assuming $e$ condition all expressions) and denoting by $F_i$ the set including $X_i$ and its child nodes,

$$E = \max_{d_1,...,d_k} \sum_{x_{k+1},...,x_n} \Pi_{i=1}^n P(x_i|x_{pa_i}, d_1, ..., d_k)u(x)$$

Applying algebraic manipulations (and denoting $d = (d_1, ..., d_k)$ and $\bar{x}_k^j = (x_k, ..., x_j)$):

$$E = \max_d \sum_{\bar{x}_{k+1}^{n-1}} \sum_{x_n} \Pi_{i=1}^n P(x_i|x_{pa_i}, d) \sum_{Q_j \in Q} f_j(x_{Q_j}).$$

We can now separate the components in the utility functions into those mentioning $X_n$, denoted by the index set $t_n$, and those not mentioning $X_n$, labeled with indexes $l_n = \{1, ..., j\} - t_n$. We separate the utility into two parts as well. We get

$$E = \max_d [\sum_{\bar{x}_{k+1}^{(n-1)}} \sum_{x_n} \Pi_{i=1}^n P(x_i|x_{pa_i}, d) \sum_{j \in l_n} f_j(x_{Q_j})$$

$$+ \sum_{\bar{x}_{k+1}^{(n-1)}} \sum_{x_n} \Pi_{i=1}^n P(x_i|x_{pa_i}, d) \sum_{j \in t_n} f_j(x_{Q_j})]$$

By migrating to the left of the summation in $X_n$ all of the elements that are not a function of $X_n$, we get:

$$= \max_d [\sum_{\bar{x}_{k+1}^{n-1}} \Pi_{X_i \in X - F_n} P(x_i|x_{pa_i}, d) \cdot$$

$$(\sum_{j \in l_n} f_j(x_{S_j})) \sum_{x_n} \Pi_{X_i \in F_n} P(x_i|x_{pa_i}, d)$$

$$+ \sum_{\bar{x}_{k+1}^{n-1}} \Pi_{X_i \in X - F_n} P(x_i|x_{pa_i}, d) \cdot$$

$$\sum_{x_n} \Pi_{X_i \in F_n} P(x_i|x_{pa_i}, d) \sum_{j \in t_n} f_j(x_{Q_j})]$$

We denote by $U_n$ the subset of variables that appear with $X_n$ in a probabilistic component, excluding $X_n$ itself, and by $W_n$ the union of variables appearing in probabilistic and utility components with $X_n$, but excluding $X_n$ itself. We define $\lambda_n$ over $U_n$ as follows ($x$ is a tuple over $U_n \cup X_n$):

$$\lambda_n(x_{U_n}|d) = \sum_{x_n} \Pi_{X_i \in F_i} P(x_i|x_{pa_i}, d). \qquad (1)$$

We define $\theta_n$ over $W_n$,

$$\theta_n(x_{W_n}|d) = \sum_{x_n} \Pi_{X_i \in F_n} P(x_i|x_{pa_i}, d) \sum_{j \in t_n} f_j(x_{Q_j})).$$

We get

$$E = \max_d \sum_{\bar{x}_{k+1}^{n-1}} \Pi_{X_i \in X - F_n} P(x_i|x_{pa_i}, d) \cdot$$

$$\lambda_n(x_{U_n}|d)[\sum_{j \in l_n} f_j(x_{S_j}) + \frac{\theta_n(x_{W_n}|d)}{\lambda_n(x_{U_n}|d)}]$$

$\theta_n$ and $\lambda_n$ compute the effect of eliminating $X_n$. When there is no evidence, $\lambda_n$ is a constant. The result is an expression that does not include $X_n$ where the product has one more matrix $\lambda_n$ and the utility components have one more element $\gamma_n = \frac{\theta_n}{\lambda_n}$. Applying this recursively yields the elimination algorithm in Figure 6. We assume that decision variables are processed last by elim-meu. Each bucket contains utility components and probability components. The $\theta_i$ are viewed as utility components. The algorithm generates the $\lambda_i$ of a bucket by multiplying all its probability components and summing over the variable's bucket. The $\theta$ of a bucket is computed as the average utility of that bucket, normalized by its $\lambda$. The resulting $\theta$ and $\lambda$ are placed into the appropriate bucket.

The maximization over the decision variables can be accomplished subsequently by using maximization as the elimination operator. Clearly maximization and summation can be interleaved to some degree, allowing more efficient orderings. The algorithm in [Kjae 93] can be viewed as a variation of elim-meu tailored to dynamic probabilistic networks. As before, the algorithm's performance can be bounded as a function of the structure of the augmented graph. The augmented graph is the moral graph augmented with arcs connecting any two variables appearing in the same utility component.

**Theorem 3.6** *Algorithm elim-meu computes the MEU of an influence diagram in $O(n \cdot exp(w * (o))$, when $w * (o)$ is the induced width along o of its augmented moral graph, and n is the number of variables.*
□

---

**Algorithm elim-meu**
**Input:** A belief network $BN = \{P_1, ..., P_n\}$; a subset of variables $D_1, ..., D_k$ are decision variables which are all root nodes; a utility function over $X$, $u(x) = \sum_j f_j(x_{Q_j})$; an ordering of the variables, $o$, in which the $D$'s appear first.
**Output:** An assignment $d_1, ..., d_k$ that maximizes the expected utility.
1. **Initialize:** Partition components into buckets, where $bucket_i$ contains all matrices whose highest variable is $X_i$. Call probability matrices $\lambda_1, ..., \lambda_j$ and utility matrices $\theta_1, ..., \theta_l$. Let $S_1, ..., S_j$ be the probability variable subsets while $Q_1, ..., Q_l$ be the utility variable subsets.
2. **Backward:** For $p \leftarrow n$ downto 1 do
for all the matrices $\lambda_1, ..., \lambda_j, \theta_1, ..., \theta_l$ in $bucket_p$ do
• If (bucket with observed variable) $bucket_p$ contains the observation $X_p = x_p$, then
assign $X_p = x_p$ to each $\lambda_i, \theta_i$ and put each resulting matrix in the appropriate bucket.
• else, $U_p \leftarrow \bigcup_{i=1}^j S_i - \{X_p\}$ and $W_p \leftarrow U_p \cup (\bigcup_{i=1}^l Q_i - \{X_p\})$. For all $U_p = u$, $\lambda_p(u) = \sum_{x_p} \Pi_i \lambda_i(x_p, u_{S_i})$ and for all $W_p = w$, $\theta_p(w) = \frac{1}{\lambda_p(w_{U_p})} \sum_{x_p} \Pi_{i=1}^j \lambda_i(x_p, w_{S_i}) \sum_{j=1}^l \theta_j(x_p, w_{Q_j})$,
Add $\theta_p$ and $\lambda_p$ to the bucket of the largest-index variable in $W_p$ and $U_p$, respectively.
3. **Forward:** Assign values in the ordering $o = D_1, ..., D_k$ using the information recorded in each bucket of the decision variable. (This can be accomplished using elimination with maximization on the rest of the decision buckets)

Figure 6: Algorithm elim-meu

---

**Algorithm elim-cond-max**
**Input:** A belief network $BN = \{P_1, ..., P_n\}$; an ordering of the variables, $o$; a subset $C$ of conditioned variables.
**Output:** The most probable assignment, given evidence $e$.
**Initialize:** $p_{max} = 0$.

1. For every combination $C = c$,
   put each conditioned variable with its new value in its bucket.
   2. $p \leftarrow elim - max(o, e, c)$ (apply elim-max when $C = c$ added as observation).
   3. $p_{max} \leftarrow \max\{p_{max}, p\}$ (keep the current maximum probability assignment).

2. Return $p_{max}$ and $argmax_x(p_{max})$.

---

Figure 7: Algorithm elim-cond-max

## 4 COMBINING ELIMINATION AND CONDITIONING

A serious drawback of elimination algorithms is that they require considerable memory to record intermediate functions. Conditioning, on the other hand, requires only linear space. Combining conditioning with elimination may reduce memory needs but still provide performance bounds.

We will demonstrate the idea on the MPE task:
$$\max_x P(x) = \max_x \Pi_i P(x_i|x_{pa_i})$$
when $x = (x_1, ..., x_n)$. Let $C$ be a subset of conditioning variables, $C \subseteq X$, $V = X - C$. Clearly,
$$\max_x P(x) = \max_{x_C} \max_{x_V} P(x_V, x_C)$$
Therefore, for every $x_C$, we compute $\max_{x_V} P(x_V, x_c)$ and a maximizing tuple
$$(x_V^{opt})(x_C) = argmax_V \{\Pi_{i=1}^n P(x_i|x_{pa_i})|C = x_C\}$$
using the elimination algorithm as before, treating the conditioned variables as observed variables. This basic step can be enumerated for all value combinations of the conditioning variables, and the tuple retaining the maximum probability will be kept. Given a particular value assignment $c$, the time and space complexity of computing the maximizing the joint probability over the rest of the variables is bounded exponentially by the induced width of the graph whose conditioning variables were deleted. We define the *conditional induced width of a graph relative to $C$ along $o$*, $w_C^*(o)$, as the induced width, along ordering $o$, of the graph after deleting the nodes in $C$. The algorithm is presented in Figure 7.

**Theorem 4.1** *Given a set of conditioning variables, the space complexity of algorithm $elim - cond - max$ is $O(n \cdot exp(w_C^*(o)))$, while its time complexity is $O(n \cdot exp(w_C^*(o) + |C|))$.*

Clearly, the algorithm can be implemented more effectively by taking advantage of shared partial assignments to the conditioned variables in $C$.

## 5 SUMMARY AND CONCLUSION

Using the bucket elimination framework, we have presented a uniform way of expressing algorithms for probabilistic reasoning. In this framework, algorithms require no special mechanism to move from singly-connected to multiply-connected networks and no conscious effort to manage the topological features of the network. For example, if algorithm elim-max is given a singly-connected network and we use an ordering having width 1 (always possible for trees), it reduces to Pearl's algorithm for that task [Pear 88]. Likewise, elim-bel is identical to Pearl's tree-propagation algorithm for belief update with the exception that it answers singleton queries. Each new query requires running elim-bel where the queried variables appear first in the ordering.

Clustering and elimination are closely related; in fact, elimination may be viewed as a directional version of tree-clustering which is "goal oriented" or "query oriented." Thus, preprocessing by elimination is geared to the particular query at hand (instead of all future queries).

The performance of elimination algorithms (as well as tree-clustering) is likely to suffer from the known difficulty with dynamic programming algorithms: exponential space (for recording the tables) and exponential time unless the problem has a small induced width. Such performance deficiencies also exist in resolution algorithms like $DR$ [DeRi 94]. One important method for reducing the space complexity is conditioning. We have shown that conditioning can be incorporated naturally on top of elimination, and that it can reduce the space complexity while still exploiting the structure (see also [Dech 96]). The combination of conditioning with elimination can be viewed as an elegant way for combining the virtues of forward and backward search.

The ideas underlying the algorithms we present are not new, and the role of dynamic programming in probabilistic reasoning has already been made explicit in the context of influence diagrams [TaSh 90]. What we provide here is a concise and uniform exposition across many tasks, which will facilitate transfer of ideas between areas of research.

From a practical point of view, bucket elimination is very easy to implement, since structure building is not separated from inference propagation. A student was able to implement elim-bel within a few weeks of being introduced to it. (The code is available by ftp.)

**Acknowledgement**

I would like to thank Irina Rish for commenting on the last version of this paper. This work was partially sup-ported by NSF grant IRI-9157636, Air Force Office of Scientific Research grant, AFOSR F49620-96-1-0224, Rockwell MICRO grant #ACM-20775 and 95-043 and Electrical Power Research Institute RP8014-06.

# References

[Arnb 85] S. Arnborg, "Efficient algorithms for combinatorial problems on graphs with bounded decomposability - A survey" *Bit* 25 (1985):2-23.

[ArPr 89] S. Arnborg and A. Proskourowski, "Linear time algorithms for NP-hard problems restricted to partial $k$-trees" *Discrete and Applied Mathematics* 23 (1989) 11-24.

[BeBr 72] U. Bertele and F. Brioschi, *Nonserial Dynamic Programming*, New York, 1972.

[Bacc 95] F. Bacchus and A. Groove, "Graphical models for preference and utility," in *Uncertainty in Artificial Intelligence (UAI-95)*, pp 3-10, 1995.

[Coop 84] G.F. Cooper, "Nestor: A computer-based medical diagnosis aid that integrates causal and probabilistic knowledge," Ph.D. dissertation, Department of Computer Science, Stanford University, 1984.

[DePe 89] R. Dechter and J. Pearl, "Tree clustering for constraint networks," *Artificial Intelligence* (1989):353-366.

[Dech 90] R. Dechter, "Constraint networks" in *Encyclopedia of Artificial Intelligence,* 2nd ed., New York, 1990.

[DeBe 95] R. Dechter and P. van Beek, "Local and global relational consistency - Summary of recent results," in *Principles and Practice of Constraint Programming (CP-95)*,, Cadec, France, 1995.

[Dech 96] R. Dechter, "Topological parameters for time-space trade-off," in *Proceedings of the 12th Conference on Uncertainty in Artificial Intelligence (UAI-96)*, 1996.

[Jens 94] F. Jensen, F. Jensen and S. Dittmer "From influence diagrams to junction trees," in *Uncertainty in Artificial Intelligence (UAI-94)*, pp. 367-373, 1994.

[DeRi 94] R. Dechter and I. Rish, "Directional resolution: The Davis-Putnam procedure, revisited," in *Principles of Knowledge Representation and Reasoning (KR-94)*, 1994.

[DaPu 60] M. Davis and H. Putnam, "A computing procedure for quantification theory," *Journal of the ACM* 7 (1960):201-216.

[Pear 88] J. Pearl, *Probabilistic Reasoning in Intelligent Systems*, 2nd ed., San Mateo, CA, 1988.

[PeRe 86] Y. Peng and J.A. Reggia "Plausability of diagnostic hypothesis", in *Proceedings (AAAI-86)*, Philadelphia, 1986, pp. 140-145.

[PeRe 89] Y. Peng and J. Reggia, "A connectionist model for diagnostic problem solving," *IEEE Transactions on Systems, Man and Cybernetics* 19 (1989): pp. .

[Sant 91] E. Santos, "On the generation of alternative explanations with implications for belief revision," *Uncertainty in Artificial Intelligence (UAI-91)*, pp. 339-347, 1991.

[Shac 86] R.D. Shachter "Evaluating influence diagrams," *Operations Research*, 34 (1986).

[Shac 88] R.D. Shachter, "Probabilistic inference and influence diagrams," *Operations Research*, 36 (1988).

[ShPe 92] R.d. Shachter and P. Peot, "Decision making using probabilistic inference methods," in *Uncertainty in Artificial Intelligence (UAI-92)*, pp. 276-283, 1992.

[Shac 90] R.D. Shachter, B. D'Ambrosio, and B.A. Del Favro, "Symbolic probabilistic inference in belief networks," *Automated Reasoning* (1990): 126-131.
In *Operations Research* Vol. 36, No.4, 198b.

[Shen 92] P.P. Shenoy, "Valuation-based systems for Bayesian decision analysis," em Operations Research, 40 (1992): 463-484.

[ShCh 91] S.E. Shimony and E. Charniack, "A new algorithm for finding MAP assignments to belief networks,". In P. Bonissone, M. Henrion, L. Kanal, and J. Lemmer ed., Uncertainty in Artificial Intelligence 6, pp. 185-193, New York, 1991.

[Spie 86] D.J. Spiegelhalter, "Probabilistic reasoning in predictive expert systems," in *Uncertainty in Artificial Intelligence*, ed. L.N. Kanal and J.F. Lemmer, pp. 47-68, Amsterdam, 1986.

[Kjae 93] U. Kjaerulff, "A computational scheme for reasoning in dynamic probabilistic networks," in *Uncertainty in Artificial Intelligence*(UAI-93), pp. 121-129, 1993.

[TaSh 90] J.A. Tatman and R.D. Shachter, "Dynamic programming and influence diagrams," *IEEE Transactions on Systems, Man, and Cybernetics* 20 (1990): 365-379.